\documentclass[runningheads]{llncs}
\usepackage{graphicx}

\usepackage{tikz}
\usepackage{comment}
\usepackage{amsmath,amssymb} 
\usepackage{color}
\usepackage{xcolor, colortbl}
\usepackage{xspace}
\usepackage{booktabs}
\usepackage{multirow}
\usepackage{pdfpages}
\usepackage[normalem]{ulem} 
\usepackage{url}

\usepackage[accsupp]{axessibility}  

\newcommand{\todo}[1]{{\color{red} \bf [TODO: #1]}}

\newcommand*{\ie}{i.e.\@\xspace}
\newcommand*{\eg}{e.g.\@\xspace}
\newcommand*{\ea}{et al.\@\xspace}

\definecolor{firstcolor}{rgb}{1, 0.6, 0.6}
\definecolor{secondcolor}{rgb}{1, 0.8, 0.6}
\definecolor{thirdcolor}{rgb}{1,1, 0.6}

\begin{document}
\pagestyle{headings}
\mainmatter
\def\ECCVSubNumber{5721}  

\title{SimpleRecon: \\ 3D Reconstruction Without 3D Convolutions} 


\titlerunning{SimpleRecon: 3D Reconstruction Without 3D Convolutions}
\authorrunning{Sayed \ea}

\author{Mohamed Sayed$^2$\thanks{Work done while at Niantic, during Mohamed's internship.}\hspace{10pt}
John Gibson$^1$\hspace{10pt}
Jamie Watson$^{1,2}$\hspace{10pt}\\
Victor Prisacariu$^{1,3}$\hspace{10pt}
Michael Firman$^1$\hspace{10pt}
Cl\'{e}ment Godard$^{4\star}$}


\institute{$^1$Niantic \hspace{12pt} $^2$UCL \hspace{12pt} $^3$University of Oxford \hspace{12pt} $^4$Google}

\maketitle

\begin{abstract}

Traditionally, 3D indoor scene reconstruction from posed images happens in two phases: per-image depth estimation, followed by depth merging and surface reconstruction. Recently, a family of methods have emerged that perform reconstruction directly in final 3D volumetric feature space. While these methods have shown impressive reconstruction results, they rely on expensive 3D convolutional layers, limiting their application in resource-constrained environments. In this work, we instead go back to the traditional route, and show how focusing on high quality multi-view depth prediction leads to highly accurate 3D reconstructions using simple off-the-shelf depth fusion. We propose a simple state-of-the-art multi-view depth estimator with two main contributions: 1) a carefully-designed 2D CNN which utilizes strong image priors alongside a plane-sweep feature volume and geometric losses, combined with 2) the integration of keyframe and geometric metadata into the cost volume which allows informed depth plane scoring. Our method achieves a significant lead over the current state-of-the-art for depth estimation and close or better for 3D reconstruction on ScanNet and 7-Scenes, yet still allows for online real-time low-memory reconstruction.  Code, models and results are available at \url{https://nianticlabs.github.io/simplerecon}


\end{abstract}

\begin{figure}[t]
\centering
  \resizebox{\textwidth}{!}{
  \newcommand{\turnheightnew}{0.25\columnwidth}

\centering

\begin{tabular}{@{\hskip 2mm}c@{\hskip 2mm}c@{\hskip 2mm}c@{\hskip 2mm}c@{}}

Input image (1/8) & DVMVS Depth~\cite{duzceker2021deepvideomvs} & \textbf{Our Depth} & GT Depth\\

\includegraphics[height=\turnheightnew]{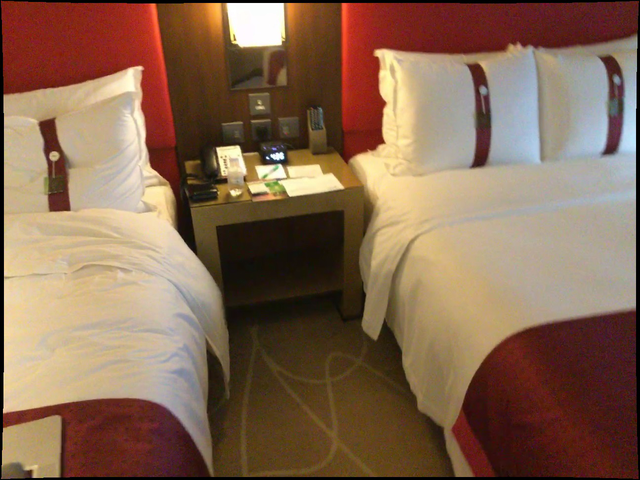} &
\includegraphics[height=\turnheightnew]{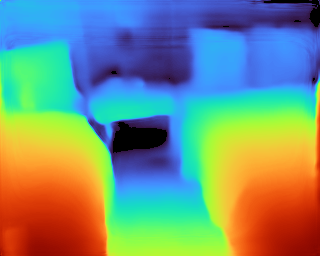} &
\includegraphics[height=\turnheightnew]{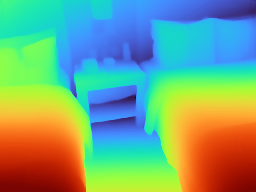} &
\includegraphics[height=\turnheightnew]{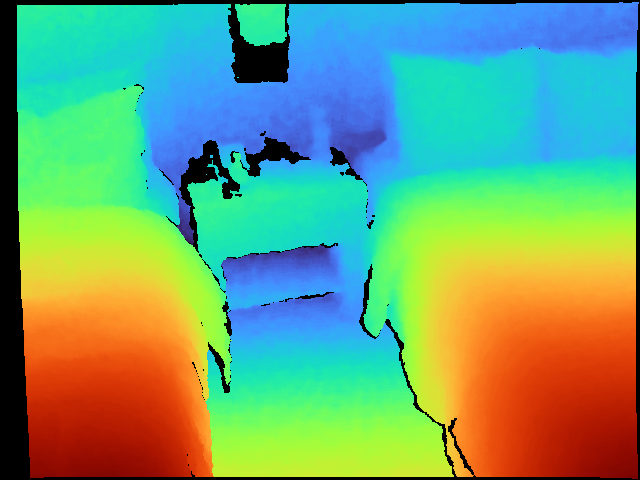} \\

\multicolumn{4}{c}{\includegraphics[width=1.0\textwidth]{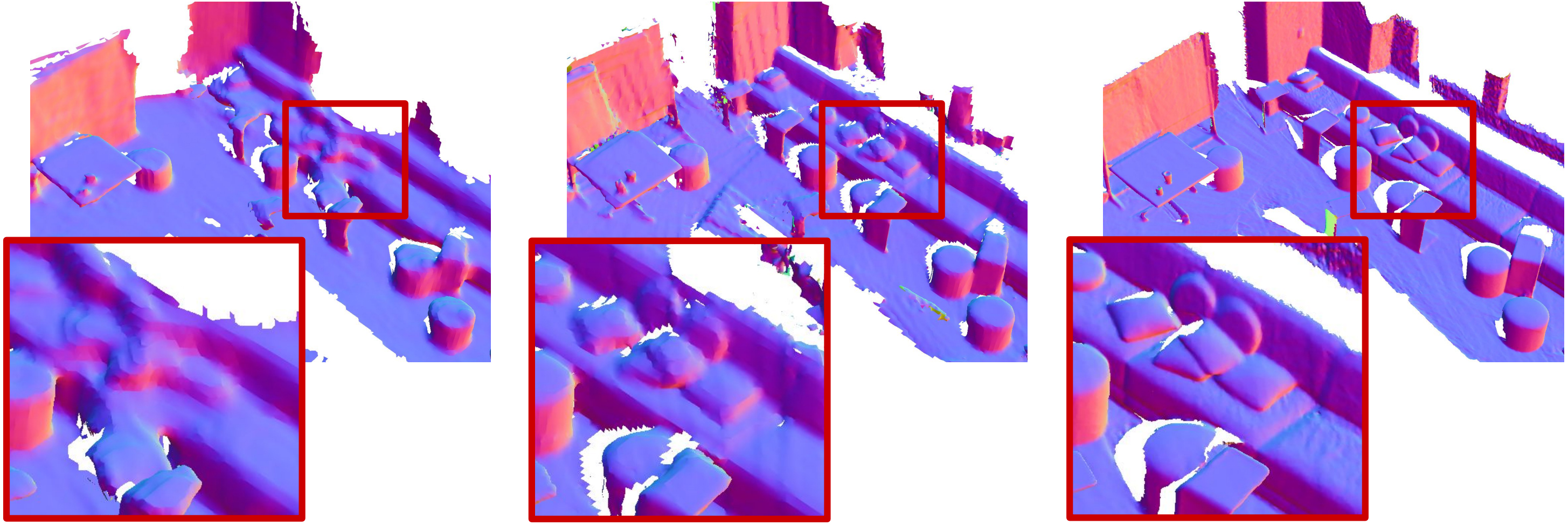}} \\
\multicolumn{4}{c}{VoRTX Mesh~\cite{stier2021vortx} \hspace{1.5cm} \textbf{Our Fused Mesh} \hspace{2.0cm} GT Mesh \hspace{2.0cm}}  \\

\end{tabular}
}
  \caption{\textbf{Qualitative preview of our method.} Our method significantly improves upon previous state-of-the-art monocular MVS methods~\cite{duzceker2021deepvideomvs} in depth prediction and matches the current volumetric state-of-the-art in full scene reconstruction~\cite{stier2021vortx}.} 
\label{fig:teaser}
\end{figure}

\section{Introduction}

Generating 3D reconstructions of a scene is a challenging problem in computer vision which is useful for tasks such as robotic navigation, autonomous driving, content placement for augmented reality  and historical preservation~\cite{newcombe2011kinectfusion,yang2020mobile3drecon}. 
Traditionally, such 3D reconstructions are generated from 2D depth maps obtained using multi-view stereo (MVS) \cite{scharstein2001stereo,drory2014sgm}, which are then fused into a 3D representation from which a surface is extracted \cite{schoenberger2016sfm,kazhdan2006poisson}. 
Recent advances in deep learning have enabled convolutional methods to outperform classical methods for depth prediction from multiple stereo images, spearheaded by GC-Net~\cite{kendall2017end} and MVSNet~\cite{yao2018mvsnet}. 
Key to these methods is the use of 3D convolutions to smooth and regularize a 4D ($C \times D \times H \times W$) cost volume, which performs well in practice but is expensive in both time and memory.
This could preclude their use on low power hardware \eg smartphones, where overall compute energy and memory are limited. 
The same is true of recent depth estimators which use LSTMs and Gaussian processes for improved depth accuracy \cite{duzceker2021deepvideomvs,hou2019multi}.


A new stream of work started by ATLAS~\cite{murez2020atlas} (and extended by \eg~\cite{sun2021neuralrecon,bozic2021transformerfusion,choe2021volumefusion}) performs the reconstruction directly in 3D space by predicting a truncated signed distance function (TSDF) from a 4D feature volume computed from the input images.
Again, these works give good results but use expensive 3D convolutions.



\begin{figure}[h]
\centering
  \resizebox{\textwidth}{!}{
  \newcommand{\turnheightnew}{0.33\columnwidth}

\centering

\begin{tabular}{@{\hskip -1mm}c@{\hskip 2mm}c@{\hskip 2mm}c@{}}

\includegraphics[height=\turnheightnew]{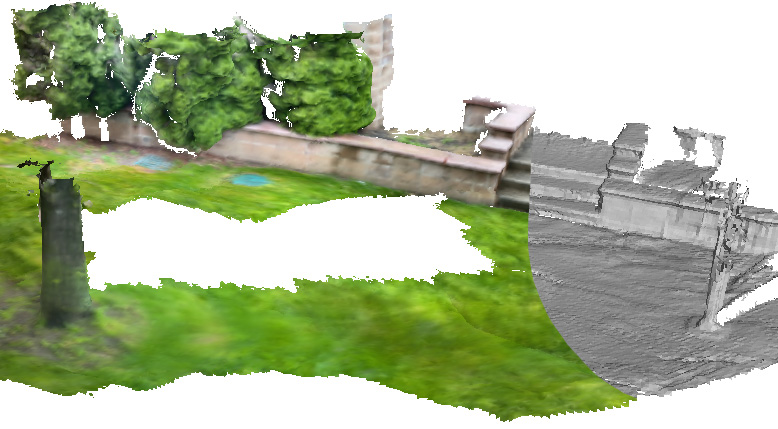} &
\includegraphics[height=\turnheightnew]{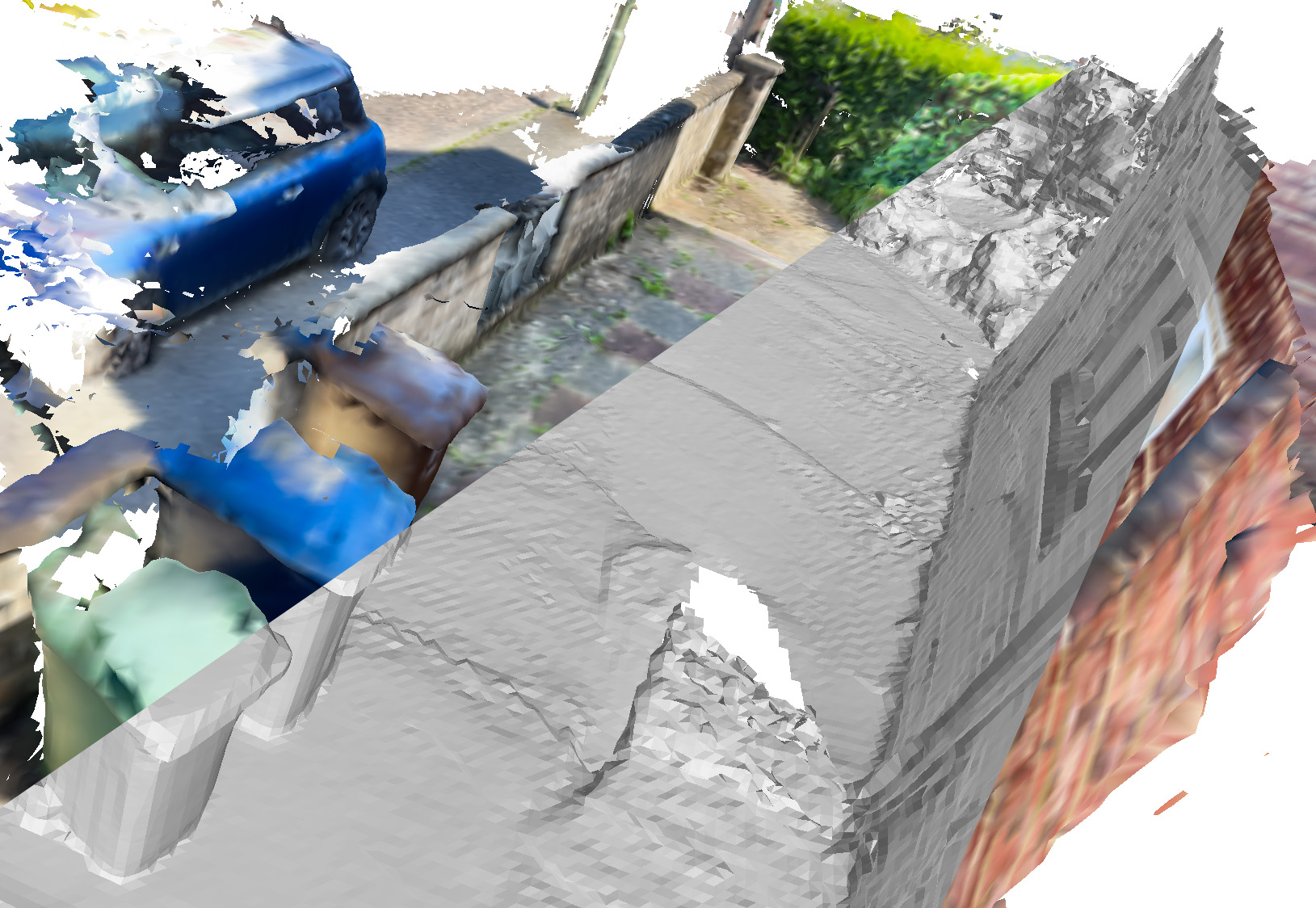} &
\includegraphics[height=\turnheightnew]{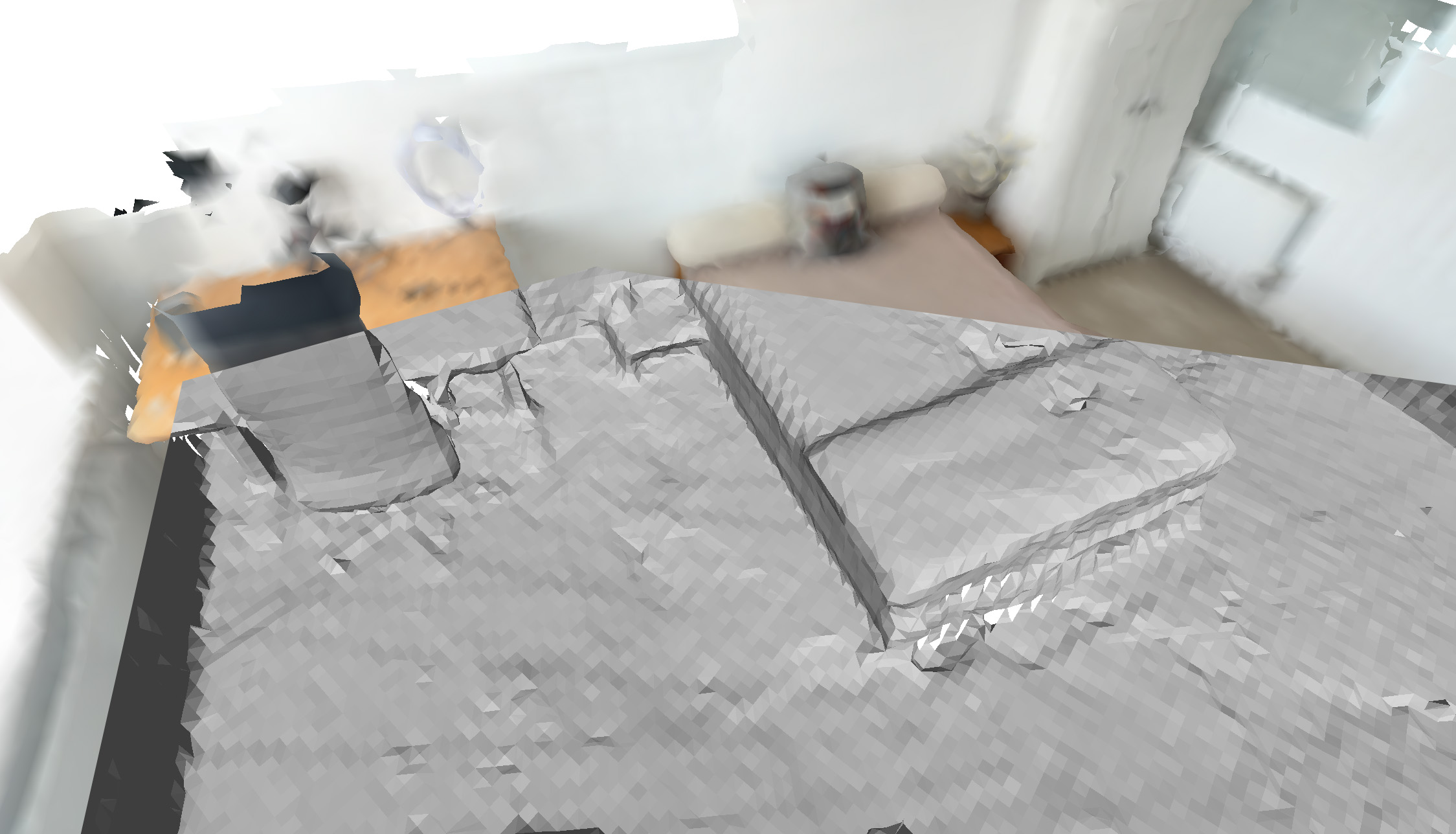} \\

\end{tabular}}
  \caption{\textbf{Fused meshes on unseen data.} Our model generalizes on unseen environments, including outdoors, captured on smartphone. See video for live reconstruction.} 
    \label{fig:unseen-data}
\end{figure}

In this paper we go \emph{back to basics}, showing that, surprisingly, it is possible to obtain state-of-the-art depth accuracy with a simple 2D CNN augmented with a cost volume.
Our method also gives competitive scores in 3D scene reconstruction using off-the-shelf TSDF fusion~\cite{newcombe2011kinectfusion}, all without expensive 3D convolutions.
Key to our method is the novel incorporation of cheaply available \emph{metadata} into the cost volume, which we show significantly improves depth and reconstruction quality.
Our main contributions are: (1) The integration of keyframe and geometric metadata into the cost volume using a multi-level perceptron (MLP), which allows informed depth plane scoring, and (2) A carefully-designed 2D CNN that utilizes strong image priors alongside a plane-sweep 3D feature volume and geometric losses.
We evaluate our `back-to-basics' method against all recent published methods on the challenging ScanNetv2~\cite{dai2017scannet} dataset on both depth estimation and 3D scene reconstruction (Sec.~\ref{sec:experiments}), and show it generalizes on 7-Scenes~\cite{glocker2013real-time} data (Table~\ref{table:depth_results}) and casually captured footage (Fig.~\ref{fig:unseen-data}).

By combining our novel cost volume metadata with principled architectural decisions that result in better depth predictions, we can avoid the computational cost associated with 3D convolutions, potentially enabling use in embedded and resource-constrained environments. 
We have released code, models and precomputed results at \url{https://nianticlabs.github.io/simplerecon}

\section{Related Work}
\label{sec:related_work}
Our method is related to prior work in \emph{stereo} depth estimation, \emph{multi-view} depth estimation, and 3D reconstruction.

\subsection{Depth from Calibrated Stereo Pairs}

Many methods for estimating depth use calibrated stereo pairs of images in order to estimate disparity, which can be translated into depth using camera parameters and the intra-axial distance between the camera positions. 
Early methods compare patches \cite{hirschmuller2007stereo,vzbontar2016stereo,mayer2015large}, similar to work in optical flow estimation \cite{fischer2015flownet}. This laid the groundwork for GCNet~\cite{kendall2017end}, which built on earlier plane-sweep stereo works \cite{collins1996space,kang2001handling} to develop now-ubiquitous cost-volume-based depth estimation.
The typical architecture is feature extraction from input images, then feature matching and reduction into a cost volume, followed by convolutional layers to output the final disparity.
Further improvements include post-processing the cost volume \cite{chang2018pyramid,Zhang2019GANet,zhang2019domaininvariant,cheng2019learning} using multiscale information, carefully designed network layers that mimic classical refinement methods, and spatial pyramid pooling.
The best results typically come from running 3D convolutions on a 4D  ($C \times D \times H \times W$) cost volume, pioneered by Chang \ea in PSMNet \cite{chang2018pyramid}; this can be very computationally expensive. A more attractive option is to create a 3D cost volume ($D \times H \times W$) by reducing along the feature dimension, meaning 2D convolutions can be used for further processing \cite{watson2021temporal,yee2020fast}; however, this typically comes at the expense of depth quality.
In this work we show how, with simple tricks and clever reduction techniques, a method with a 3D cost volume can outperform existing 4D cost volume methods for both depth estimation and 3D scene reconstruction.


\subsection{Multi-view Stereo Depth}

Multi-View Stereo (MVS) is a more general problem which aims to estimate depth at a \textit{reference} viewpoint using one or more additional \textit{source} viewpoints captured from arbitrary locations. Knowledge of camera intrinsics and extrinsics for both reference and source views are generally assumed, but can also be estimated offline using \eg structure-from-motion \cite{schoenberger2016sfm} or on-line using inertial and camera tracking, like that provided by ARKit or ARCore.

Classical MVS methods typically use patch matching with photometric consistency to estimate a depth map followed by depth fusion and refinement~\cite{furukawa2015multi,schoenberger2016mvs}.
In contrast, early learning-based methods backprojected dense image features from multiple viewpoints into 3D volumes representing the entire scene and then predicted voxel occupancy or surface probability from a fused 3D volume \cite{ji2017surfacenet,kar2017learning}.
Recent methods, inspired by binocular stereo matching techniques, combine these approaches, performing epipolar-geometry-consistent matching on image pixels (\eg in MVDepthNet \cite{wang2018mvdepthnet} and DeepMVS~\cite{huang2018DeepMVS}), or extracted features (\eg in DPSNet~\cite{im2019dpsnet} and DeepVideoMVS~\cite{duzceker2021deepvideomvs}) to produce a matching cost volume. The cost volume can optionally be reduced using a dot product \cite{duzceker2021deepvideomvs} or mean absolute difference \cite{watson2021temporal,newcombe2011dtam}, and then processed using convolutional layers. Further works incorporate additional scene information to regularize the cost volume and refine the final output by using reference image features~\cite{yao2018mvsnet}, by taking into account occlusions~\cite{long2020occlusion} and moving objects~\cite{wimbauer2020monorec}, or with a Gaussian process prior~\cite{hou2019multi}. Others have proposed methods to combine multiple reference views, \eg by pooling in DeepMVS~\cite{huang2018DeepMVS} or averaging the feature volumes in DPSNet~\cite{im2019dpsnet,long2020occlusion}. Aside from use of keyframe image values and features in a cost volume, depth-estimation approaches have used temporal information in varying ways, such as  LSTMs to fuse volumes over multiple frames~\cite{tananaev2018temporally,duzceker2021deepvideomvs,patil2020dont}, or by a test-time optimization of reprojection error \cite{casser2018depth,chen2019self,luo2020consistent,mccraith2020monocular,shu2020feature,kuznietsov2021comoda}. 
However, all these approaches use only the color image as inputs, discarding additional information such as viewing direction and relative pose estimation after the cost volume is computed. 
In this work, we extend the matching cost volume into a matching feature volume, which uses readily-available metadata to produce higher-quality depth maps.

\subsection{3D Scene Reconstruction from Posed Views}




Classical methods for creating dense 3D reconstructions from images typically compute dense depth per-view, such as~\cite{schoenberger2016mvs}, followed by a surface reconstruction such as Delaunay triangulation~\cite{lee1980two} or Poisson surface reconstruction~\cite{kazhdan2006poisson}. 
The seminal work Kinect Fusion~\cite{newcombe2011kinectfusion} demonstrated \emph{real-time} 3D scene reconstruction from depth-maps using a volumetric truncated signed distance field (TSDF) representation~\cite{curless1996volumetric}, from which a mesh can be obtained using marching cubes~\cite{lorensen1987marching}. A family of methods improved on it, allowing it to work more efficiently on larger scenes~\cite{whelan2012kintinuous,niessner2013real,kahler2015hierarchical,prisacariu2017infinitam}, to handle moving objects \cite{scona2018staticfusion,runz2018maskfusion}, or to perform loop closure \cite{whelan2015elasticfusion}, all of which solidified TSDF fusion as a key component of real-time mapping.

Recent deep learning methods forego depth estimation, instead extracting 2D image features from keyframes and backprojecting these features into 3D space to produce a 4D feature volume~\cite{sitzmann2018deepvoxels}. In ATLAS~\cite{murez2020atlas}, 3D convolutions on such a feature volume are used to regress a TSDF for the scene, which significantly improved reconstruction quality over the then-state-of-the-art method of learning-based MVS followed by traditional TSDF fusion~\cite{newcombe2011kinectfusion}. NeuralRecon~\cite{sun2021neuralrecon} extended this to refine the TSDF in a coarse-to-fine manner using recurrent layers, while TransformerFusion~\cite{bozic2021transformerfusion} and VoRTX~\cite{stier2021vortx} further improved performance using transformers~\cite{vaswani2017attention} to learn feature matching.
Recently methods proposed combining volumetric reasoning -- via a 3D encoder-decoder -- with MVS reconstruction; either iteratively in the case of 3DVNet~\cite{rich20213dvnet} or using pose-invariant 3D convolutional layers in VolumeFusion~\cite{choe2021volumefusion}.

Although these methods produce high-quality reconstructions, the use of 3D convolutions, transformers, or recurrent layers makes them computationally expensive and memory-intensive. Furthermore, they predict the whole scene TSDF at once, making real-time use impossible, or rely on complex sparsification \cite{sun2021neuralrecon} or attention mechanisms \cite{bozic2021transformerfusion} to allow for progressive updates. In contrast, we take a simpler approach: by focusing on predicting high-quality depth maps, we are able to use efficient off-the-shelf TSDF fusion methods such as Infinitam~\cite{prisacariu2017infinitam}. This allows our method to achieve real-time and progressive 3D reconstructions at low compute and memory footprints, with accuracy competitive with volumetric methods but without the use of 3D convolutions.


\section{Method}
\label{sec:method}
We take as input a reference image $\mathbf{I}^0$, a set of source images $\mathbf{I}^{n\in \{1,\ldots,N-1\}}$, as well as their intrinsics and relative camera poses.
At training time we also assume access to a ground truth depth map $\mathbf{D}^\text{gt}$ aligned with each RGB image; at test time our aim is to predict dense depth maps ${\hat{\mathbf{D}}}$ for each reference image.


\begin{figure}[bt]
    \centering
    \includegraphics[width=0.99\textwidth, trim=0 3.5cm 0 0, clip, , page=2]{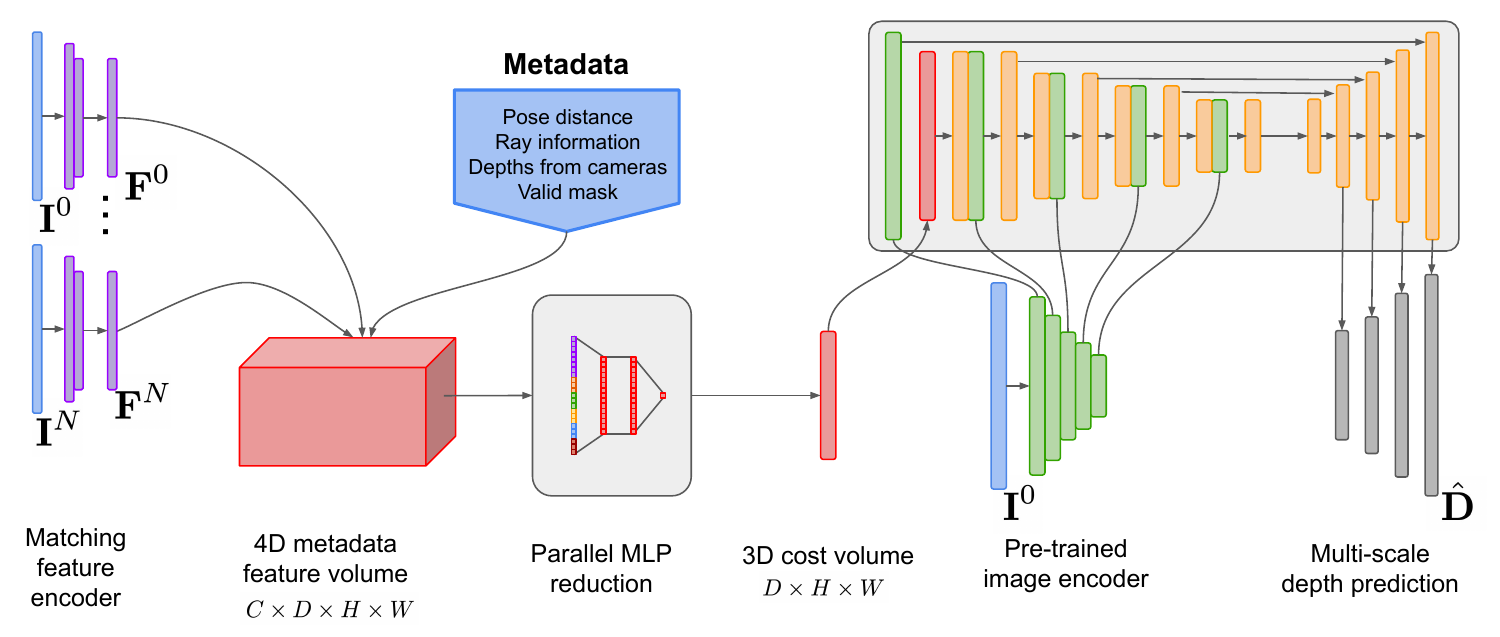}
    \caption{\textbf{Overview of our method.} Our key contribution is the injection of cheaply-available \emph{metadata} into the feature volume. Each volumetric cell is then reduced in parallel with an MLP into a feature map before input into a 2D encoder-decoder~\cite{zhou2018unet++}.}
    \label{fig:method-overview}
\end{figure}

\subsection{Method Overview}

Our depth estimation model sits at the intersection of monocular depth estimation \cite{eigen2014depth,godard2017unsupervised} and MVS via plane sweep \cite{collins1996space}. 
We augment a depth prediction encoder-decoder architecture with a cost volume; see Figure~\ref{fig:method-overview}. 
Our image encoder extracts matching features from the reference and source images for input to a cost volume. The output of the cost volume is processed using a 2D convolutional encoder-decoder network, augmented with image level features extracted using a separate pretrained image encoder.

Our key insight is to inject readily available \emph{metadata} into the cost volume alongside the typical deep image features, allowing the network access to useful information such as geometric and relative camera pose information. 
Figure~\ref{fig:metadata} shows in detail the construction of our feature volume. 
By incorporating this previously unexploited information, our model is able to significantly outperform previous methods on depth prediction without the need for costly 4D cost volume reductions~\cite{kendall2017end,yao2018mvsnet}, complex temporal fusion~\cite{duzceker2021deepvideomvs}, or Gaussian processes~\cite{hou2019multi}.

We first describe our novel metadata component and explain how it is incorporated into the network (Section~\ref{sec:metadata}).
We then set out our network architecture and losses (Sections~\ref{sec:network} and \ref{sec:loss}), giving best practices  for depth estimation.





\subsection{Improving the Cost Volume with Metadata}
\label{sec:metadata}


In traditional stereo techniques, there exists important information which is typically ignored. In this work, we incorporate readily available \emph{metadata} into the cost volume, allowing our network to aggregate information across views in an informed manner.
This can be done both \emph{explicitly} via appending extra feature channels and \emph{implicitly} via enforcing specific feature ordering.

We propose injecting metadata into our network by augmenting image-level features inside the cost volume with additional metadata channels. These channels encode information about the 3D relationship between the images used to build the cost volume, allowing our network to better reason about the relative importance of each source image for estimating depth for a particular pixel. 

\begin{figure}[t]
    \centering
    \includegraphics[width=0.99\textwidth,trim=0 7cm 0 0, clip, page=1]{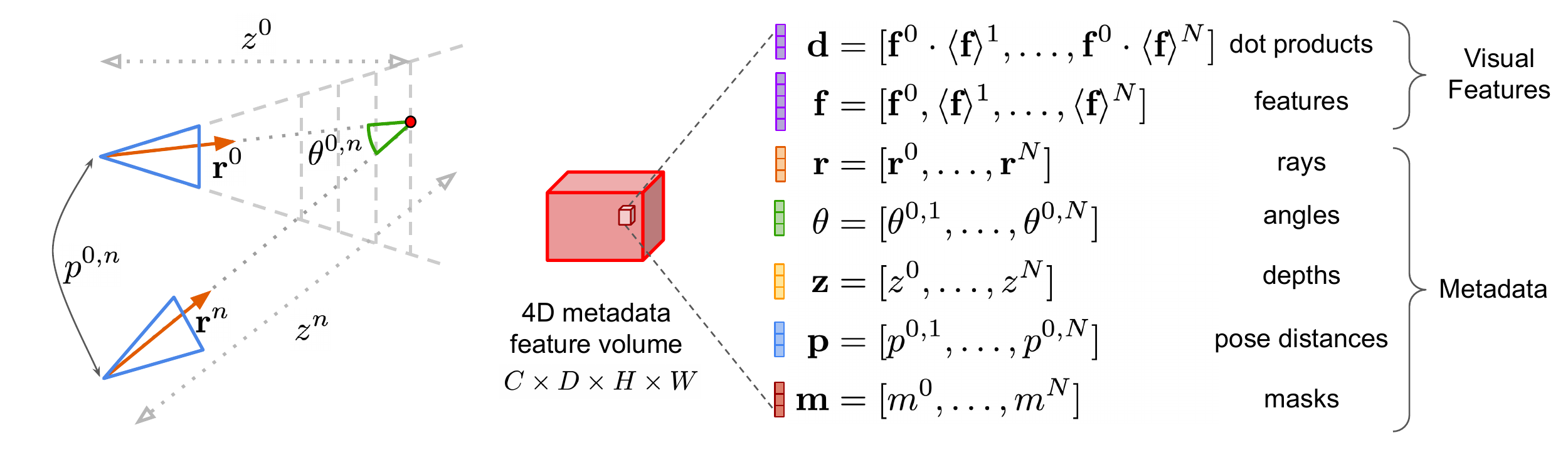}
    \caption{\textbf{Metadata insertion.} Typical MVS systems predict depth from warped \emph{features} or differences between features \eg dot products. We additionally include cheaply-available \emph{metadata} for improved performance. Indices $(k,i,j)$ are omitted for clarity.}
    \label{fig:metadata}
\end{figure}

Our cost volume is therefore a 4D tensor of dimension $C \times D \times H \times W$, where for each spatial location $(k,i, j)$, $k$ is the depth plane index, we have a $C$ dimensional feature vector.
This vector comprises reference image features $\textbf{f}^0_{k, i, j}$ and a set of warped source image features $\langle\textbf{f}\rangle^n_{k, i, j}$ for $n \in [1, N]$, where $\langle\;\rangle$ indicates that the features are  perspective-warped into the reference camera frame, along with the following metadata components:

\begin{description}
    \setlength\itemsep{0.15em}

    \item[Feature dot product] --- The dot product between reference image features and warped source image features, \ie $\textbf{f}^0 \cdot \langle\textbf{f}\rangle^n$. This is commonly used as the \textit{only} matching affinity in the cost volume.
    
    \item[Ray directions $\mathbf{r}^0_{k, i, j} ~ \text{and} ~ \mathbf{r}^n_{k, i, j} \in \mathbb{R}^3$] --- 
    The normalized direction to the 3D location of a point $(k,i, j)$ in the plane sweep from the camera origins.
    
        
    \item[Reference plane depth $z^0_{k,i,j}$] --- 
    The perpendicular depth from the reference camera to the point at position $k,i,j$ in the cost volume.
    
    \item[Reference frame reprojected depths $z^n_{k,i,j}$] ---
    The perpendicular depth of the 3D point at position $k,i,j$ in the cost volume to source camera $n$. 
    
    \item[Relative ray angles $\theta^{0, n}$] --- The angle between $\mathbf{r}^0_{k, i, j}$ and $\mathbf{r}^n_{k, i, j}$.
    
    \item[Relative pose distance $p^{0,n}$] --- A measure of the relative pose distance between the pose of the reference camera and each source frame \cite{duzceker2021deepvideomvs}:
    \begin{align} \label{eqn:pose_dist}
    p^{0,n} =
    \sqrt{||\mathbf{t}^{0,n}|| + \frac{2}{3}\text{tr}
    (\mathbb{I}-\textbf{R}^{0,n})}
    \end{align}
    
    \item[Depth validity masks $m^n_{k,i,j}$] --- This binary mask indicates if point $(k,i,j)$ in the cost volume projects in front of the source camera $n$ or not.
    
\end{description}


An overview of these features is given in Fig~\ref{fig:metadata}.
Each resulting $\textbf{f}_{k, i, j}$ is processed by a simple multi layer perceptron (MLP), outputting a single scalar value for each location $(k,i,j)$. This scalar can be thought of an initial estimate of the likelihood that the depth of pixel $i, j$ is equal to the $k$th depth plane.

\subsubsection{Metadata motivation ---}
We argue that by appending metadata-derived features into our cost volume, the MLP can learn to correctly weigh the contribution of each source frame at each pixel location.
Consider for instance the pose distance $p^{s,n}$; it is clear that for depths farther from the camera, the matching features from source frames with a greater baseline would be more informative. 
Similarly, ray information can be useful for reasoning about occlusions; if features from the reference frame disagree with those from a source frame but there is a large angle between camera rays, then this could be explained by an occlusion rather than incorrect depth. 
Depth validity masks can help the network to know whether to trust features from source camera $n$ at $(k,i, j)$,
By allowing our network access to this kind of information, we give it the ability to conduct such geometric reasoning when aggregating information from multiple source frames.

\subsubsection{Implicit metadata incorporation ---}
In addition to explicitly providing metadata as extra features, we also propose \emph{implicitly} encoding metadata via feature ordering. This is made possible by the inherent order dependence of MLP networks, which we exploit by choosing the ordering in which we stack our source features $\textbf{f}^n$. 
We advocate ordering $\textbf{f}^n$ by frame pose distance $p^{s,n}$, a measure shown by \cite{duzceker2021deepvideomvs,hou2019multi} to be effective for optimal keyframe selection. This ordering allows the MLP to learn a prior on pose distance and feature relevance.

Our experiments show that by including metadata in our network, both \emph{explicitly} via extra feature channels and \emph{implicitly} via feature ordering, we can obtain a significant boost to depth estimation accuracy, bringing with it improved 3D reconstruction quality; see Table \ref{table:ablations}. 
Whilst previous works have included tensors related to camera intrinsics \cite{facil2019cam} and extrinsics \cite{zhao2021camera} for monocular depth estimation, we believe that our use of metadata is a novel innovation for multi-view-stereo depth estimation. 

\subsection{Network Architecture Design}
\label{sec:network}
Our network is based on a 2D convolutional encoder-decoder architecture similar to prior works such as~\cite{duzceker2021deepvideomvs,watson2021temporal}. 
When constructing such networks, we find that there are important design choices which can give significant improvements to depth prediction accuracy. We specifically aim to keep the overall architecture simple, avoiding complex structures such as LSTMs~\cite{duzceker2021deepvideomvs} or GPs~\cite{hou2019multi}, and making our baseline model lightweight and interpretable.

\vspace{5pt}
\noindent\textbf{Baseline cost volume fusion ---} 
While RNN-based temporal fusion methods are often used~\cite{sun2021neuralrecon,duzceker2021deepvideomvs}, they significantly increase the complexity of the system. We instead make our baseline cost-volume fusion as simple as possible and find that simply summing the dot-product matching costs between the reference view and each source view leads to results competitive with state-of-the-art depth estimation techniques, as shown in Table~\ref{table:depth_results} with the heading ``no metadata".

\vspace{5pt}
\noindent\textbf{Image encoder and feature matching encoder ---}
Prior depth estimation works have shown the impact of more powerful image encoders for the task of depth estimation, both in monocular~\cite{godard2019digging,watson2019depthints,Ranftl2020} and multi-view estimation~\cite{watson2021temporal}. DeepVideoMVS~\cite{duzceker2021deepvideomvs} make use of an MnasNet~\cite{tan2019mnasnet} as their image encoder, chosen for its relatively low latency. We propose instead utilizing a still-small but more powerful EfficientNetv2 S encoder~\cite{tan2021efficientnetv2}, the smallest of its family. While this does come with a cost of increased parameter count and 10\% slower execution, it yields a sizeable improvement to depth estimation accuracy, especially for precise metrics such as Sq Rel and $\delta < 1.05$. See Table \ref{table:ablations} for full results.

For producing matching feature maps, we use the first two blocks from ResNet18~\cite{he2016deep} for efficiency, we experimented with FPN~\cite{lin2017feature} following~\cite{duzceker2021deepvideomvs}, which slightly improved accuracy at the expense of a 50\% slower overall run-time.

\vspace{5pt}
\noindent\textbf{Fuse multi-scale image features into the cost volume encoder ---}
In 2D CNN based deep stereo and multi-view stereo, image features are typically combined with the output of the cost volume at a single scale \cite{watson2021temporal,Liang2018Learning}.

More recently, DeepVideoMVS \cite{duzceker2021deepvideomvs} instead proposed concatenating deep image features at \emph{multiple} scales, adding skip connections between the image encoder and cost volume encoder at all resolutions. Whilst this has been shown to be helpful for their LSTM-based fusion network, we find that it is similarly important for our architecture.

\vspace{5pt}
\noindent\textbf{Number of source images ---}
While other methods show diminishing returns as additional source frames are added~\cite{duzceker2021deepvideomvs}, our method is better able to incorporate this additional information and displays increased performance with up to 8 views. We posit that incorporating additional metadata for each frame allows the network to make a more informed decision about the relative weightings of each frame's features when inferring the final cost. In contrast, methods such as MVDepthNet~\cite{wang2018mvdepthnet}, MVSNet~\cite{yao2018mvsnet}, ManyDepth~\cite{watson2021temporal} and ATLAS~\cite{murez2020atlas} give each frame equal weight during a update, thus potentially overwhelming the most useful information with lower-quality features.

\subsection{Loss}
\label{sec:loss}

We supervise our training using a combination of geometric losses, inspired by recent MVS methods~\cite{duzceker2021deepvideomvs,eigen2014depth,yao2018mvsnet,huang2018DeepMVS,yao2018mvsnet} as well as monocular depth estimation techniques~\cite{godard2019digging,li2018megadepth,yin2021learning,Ranftl2020}. We find that careful choice of loss function is required for best performance, and that supervising intermediate predictions at lower output scales substantially improves results.

\vspace{5pt}
\noindent{\textbf{Depth regression loss ---}}
We follow \cite{eigen2014depth} and densely supervise predictions using log-depth, but use an absolute error on log depth for each scale $s$,
\begin{align}
    \mathcal{L}_{\text{depth}} &= \frac{1}{HW}\sum_{s=1}^{4}\sum_{i,j} \frac{1}{s^2} | \uparrow_{gt}\! \log \hat{\mathbf{D}}^s_{i,j}  - \log \mathbf{D}^\text{gt}_{i,j} |,
\end{align}
where we upsample each lower scale depth using nearest neighbor upsampling~\cite{duzceker2021deepvideomvs} to the highest scale we predict at with the $\uparrow_{gt}$ operator.
We average this loss per pixel, per scale and per batch. 
Our experiments found this loss to perform better than the scale-invariant formulation of Eigen \ea~\cite{eigen2014depth,bhat2021adabins}, while producing much sharper depth boundaries, resulting in higher fused reconstruction quality.

\vspace{5pt}
\noindent{\textbf{Multi-scale gradient and normal losses  ---}}
We follow \cite{li2018megadepth,yin2021learning,Ranftl2020} and use a multi-scale gradient loss on our highest resolution network output
\begin{align}
    \mathcal{L}_{\text{grad}} &= \frac{1}{HW}\sum_{s=1}^{4}\sum_{i,j} | \nabla\!\downarrow_s\!\hat{\mathbf{D}}_{i,j}  - \nabla\! \downarrow_s \! \mathbf{D}^\text{gt}_{i,j} |,
\end{align}
where $\nabla$ is first order spatial gradients and $\downarrow_s$ represents downsampling to scale $s$.
Inspired by \cite{yin2019enforcing} we also use a simplified normal loss, where $\mathbf{N}$ is the normal map computed using the depth and intrinsics (see supp. mat. for details),
\begin{align}
    \mathcal{L}_{\text{normals}} &= \frac{1}{2HW} \sum_{i,j} 1 - \hat{\mathbf{N}}_{i,j} \cdot \mathbf{N}_{i,j}.
\end{align}

\vspace{5pt}
\noindent{\textbf{Multi-view depth regression loss ---}}
We use ground-truth depth maps for each source view as additional supervision by projecting predicted depth $\hat{D}$ into each source view and averaging absolute error on log depth over all valid points,
\begin{align}
    \mathcal{L}_{\text{mv}} &= \frac{1}{NHW}\sum_{n} \sum_{i,j} |\log \hat{\mathbf{D}}^{0 \rightarrow n}_{i,j}  
    - \log \mathbf{D}_{n,i,j}^{\text{gt}} | 
\end{align}
where $\hat{\mathbf{D}}^{0 \rightarrow n}$ is the depth predicted for the reference image of index $0$, projected into source view $n$. This is similar in concept to the depth regression loss above, but for simplicity is applied only on the final output scale.

\vspace{5pt}
\noindent{\textbf{Total loss ---}}
Overall our total loss is:
\begin{align}
    \mathcal{L} &= \mathcal{L}_{\text{depth}} + \alpha_{\text{grad}}\mathcal{L}_{\text{grad}} + \alpha_{\text{normals}}\mathcal{L}_{\text{normals}} + \alpha_{\text{mv}}\mathcal{L}_{\text{mv}},
\end{align}
with $\alpha_{\text{grad}}=1.0=\alpha_{\text{normals}}=1.0$, and $\alpha_{\text{mv}}=0.2$, chosen experimentally using the validation set.

\subsection{Implementation Details}

We implemented the method using PyTorch~\cite{paszke2019pytorch,falcon2019pytorch,rw2019timm} and we use an EfficientNetV2 S backbone~\cite{tan2021efficientnetv2}, with a decoder similar to UNet++~\cite{zhou2018unet++}, and use the first 2 blocks of ResNet18 (R18) for matching feature extraction. Please see supplementary material for a detailed architecture description.
We train with the AdamW optimizer~\cite{loshchilov2017decoupled} for 100k steps -- approximately 9 epochs -- with a weight decay of $10^{-4}$, and a learning rate of $10^{-4}$ for 70k steps, $10^{-5}$ until 80k, then dropped to $10^{-6}$ for remainder, which takes 36 hours on two 40GB A100 GPUs. Models with the lowest validation loss are used for evaluation.
We resize images to $512\times384$ and predict depth at half that resolution. When training, random color augmentations to brightness, contrast, saturation, and hue are applied per image using TorchVision~\cite{marcel2010torchvision} with $\delta=0.2$ for all parameters, and horizontal flips with a probability of 50\%. Keyframes are selected following DeepVideoMVS~\cite{duzceker2021deepvideomvs}.



\section{Experiments}
\label{sec:experiments}
We train and evaluate our method on the 3D scene reconstruction dataset ScanNetv2~\cite{dai2017scannet}, which comprises 1,201 training, 312 validation, and 100 testing scans of indoor scenes, all captured with a handheld RGBD sensor. We also evaluate our ScanNetv2 models without fine-tuning on 7-Scenes~\cite{shotton2013scene} using \cite{duzceker2021deepvideomvs}'s test split.


\begin{table*}[tb]
    \begin{center}
    \setlength{\tabcolsep}{0.3em} 
    \resizebox{0.99\columnwidth}{!}{  
        \begin{tabular}{lccccc|ccccc}
            \toprule
            & \multicolumn{5}{c}{\textbf{ScanNetv2}} & \multicolumn{5}{c}{\textbf{7Scenes}} \\
            \cmidrule(lr){2-6}\cmidrule(lr){7-11}
            & Abs Diff$\downarrow$  & Abs Rel$\downarrow$ & Sq Rel$\downarrow$ & $\delta < 1.05\uparrow$ &  $\delta < 1.25\uparrow$ & Abs Diff$\downarrow$  & Abs Rel$\downarrow$ & Sq Rel$\downarrow$ & $\delta < 1.05\uparrow$ &  $\delta < 1.25\uparrow$ \\
            \midrule
            DPSNet ~\cite{im2019dpsnet} & 0.1552 & 0.0795 & 0.0299 & 49.36 & 93.27 & 0.1966 & 0.1147 & 0.0550 & 38.81 & 87.07\\
            MVDepthNet ~\cite{wang2018mvdepthnet}  & 0.1648 & 0.0848 & 0.0343 & 46.71 & 92.77 & 0.2009 & 0.1161 & 0.0623 & 38.81 & 87.70\\
            DELTAS~\cite{sinha2020deltas} & 0.1497 & 0.0786 & 0.0276 & 48.64 & 93.78 & 0.1915 & 0.1140 & 0.0490 & 36.36 & 88.13\\
            GPMVS ~\cite{hou2019multi}  & 0.1494 & 0.0757 & 0.0292 & 51.04 & 93.96 & 0.1739 & 0.1003 & 0.0462 & 42.71 & 90.32\\
            DeepVideoMVS, fusion~\cite{duzceker2021deepvideomvs}* &
             \cellcolor{thirdcolor}0.1186 & \cellcolor{thirdcolor}0.0583 & \cellcolor{thirdcolor}0.0190 & \cellcolor{thirdcolor}60.20 & \cellcolor{thirdcolor}96.76 & \cellcolor{thirdcolor}0.1448 & \cellcolor{thirdcolor}0.0828 & \cellcolor{thirdcolor}0.0335 &  \cellcolor{thirdcolor}47.96 & \cellcolor{thirdcolor}93.79\\
            \textbf{Ours} (no metadata) & \cellcolor{secondcolor}0.0941 & \cellcolor{secondcolor}0.0467 & \cellcolor{secondcolor}0.0139 & \cellcolor{secondcolor}70.48 & \cellcolor{secondcolor}97.84 & \cellcolor{secondcolor}0.1105 & \cellcolor{secondcolor}0.0617 & \cellcolor{secondcolor}0.0175 &  \cellcolor{secondcolor}57.30 &  \cellcolor{secondcolor}97.02\\
            \textbf{Ours} & \cellcolor{firstcolor}0.0885 & \cellcolor{firstcolor}0.0434 & \cellcolor{firstcolor}0.0125 &  \cellcolor{firstcolor}73.16 &  \cellcolor{firstcolor}98.09 & \cellcolor{firstcolor}0.1045 & \cellcolor{firstcolor}0.0575 & \cellcolor{firstcolor}0.0153  & \cellcolor{firstcolor}59.78 & \cellcolor{firstcolor}97.38\\
            \bottomrule
        \end{tabular}
    }
    \end{center}
    \caption{\textbf{Depth evaluation.} 
    For each metric, the best-performing method is marked in red, the second-best in orange, and the third-best in yellow. Results for previous methods were taken from~\cite{duzceker2021deepvideomvs}, or evaluated for each method using their keyframes. 
    *We boosted \cite{duzceker2021deepvideomvs}’s scores by using three inference frames instead of two. \cite{duzceker2021deepvideomvs} also use a custom 90/10 split; we show SimpleRecon results using this in the supplementary.}
    \label{table:depth_results}
\end{table*}

\subsection{Depth Estimation}
In Table~\ref{table:depth_results}, we evaluate the depth predictions from our network using the metrics established in Eigen \ea \cite{eigen2014depth}. 
We also introduce a tighter threshold tolerance $\delta < 1.05$ to differentiate between high quality models.
We directly compare to previously published results, including DeepVideoMVS \cite{duzceker2021deepvideomvs}, on ScanNetv2 and 7-Scenes (Table~\ref{table:depth_results}).
 
\begin{figure}[!ht]
  \centering
  \resizebox{\textwidth}{!}{
  \newcommand{\turnheightnew}{0.195\columnwidth}
\newcommand{\colone}{scene0743_00_000800}
\newcommand{\coltwo}{scene0734_00_000320}
\newcommand{\colthree}{scene0717_00_000080}
\newcommand{\colfour}{scene0721_00_002720}
\newcommand{\colfive}{scene0713_00_000440}

\centering

\begin{tabular}{@{\hskip -5mm}c@{\hskip 1mm}c@{\hskip 1mm}c@{\hskip 1mm}c@{\hskip 1mm}c@{\hskip 1mm}c@{}}

{\rotatebox{90}{\hspace{2mm}\small Source frame}} &
\includegraphics[height=\turnheightnew]{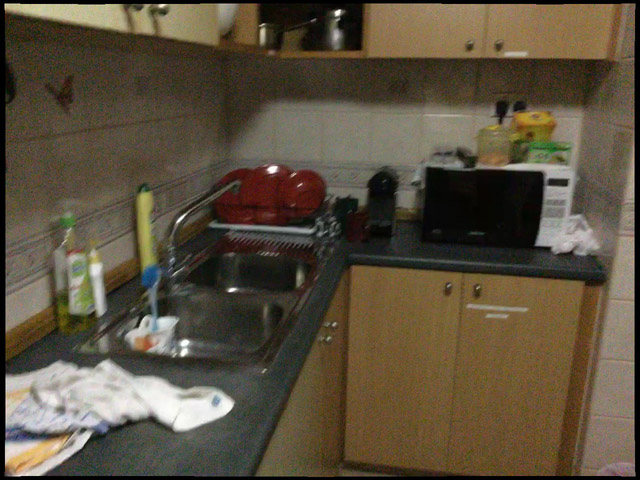} &
\includegraphics[height=\turnheightnew]{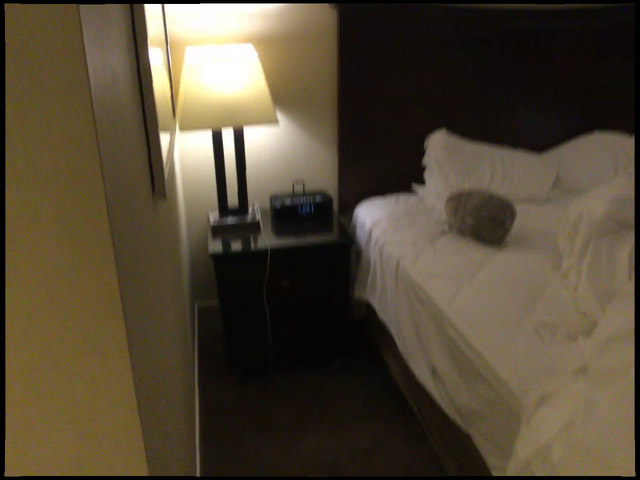} &
\includegraphics[height=\turnheightnew]{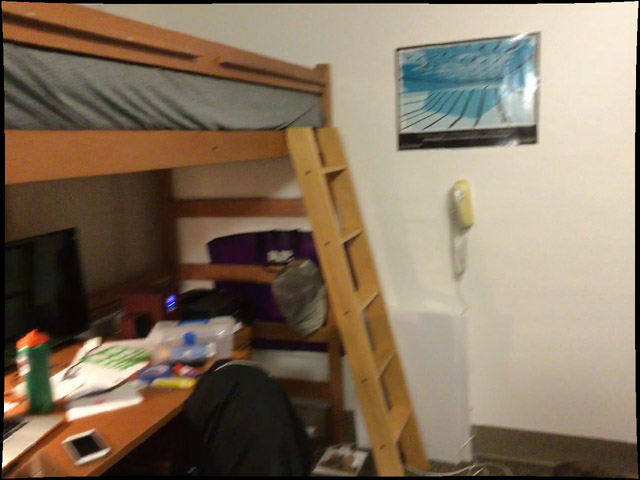} &
\includegraphics[height=\turnheightnew]{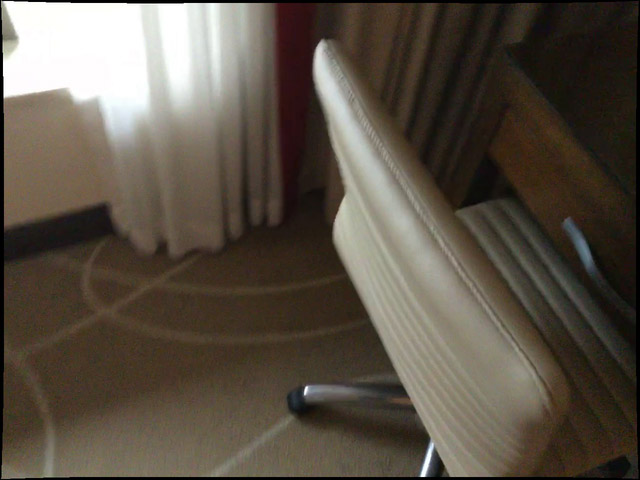} &
\includegraphics[height=\turnheightnew]{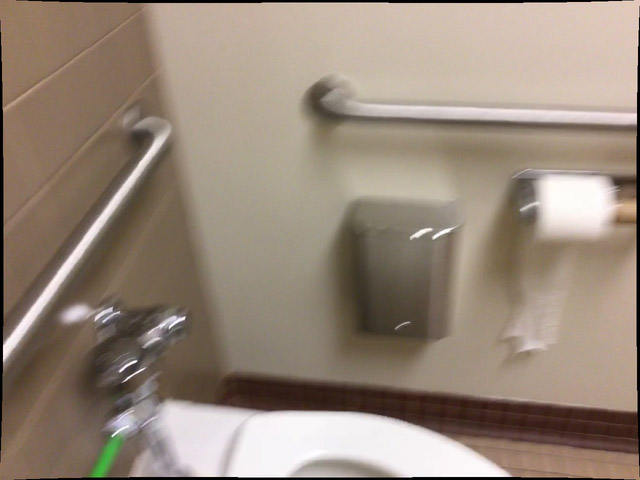} \\

{\rotatebox{90}{\hspace{1mm}{\small ESTDepth~\cite{long2021multi}}}} &
\includegraphics[height=\turnheightnew]{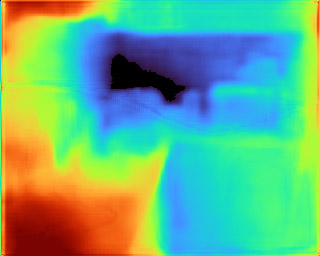} &
\includegraphics[height=\turnheightnew]{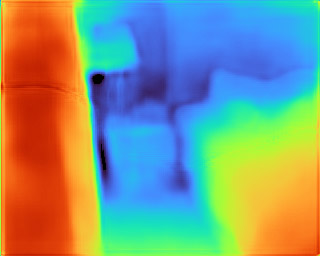} &
\includegraphics[height=\turnheightnew]{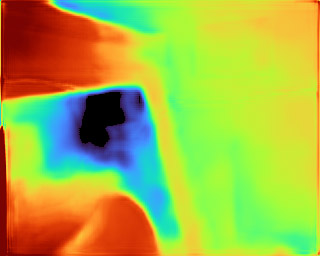} &
\includegraphics[height=\turnheightnew]{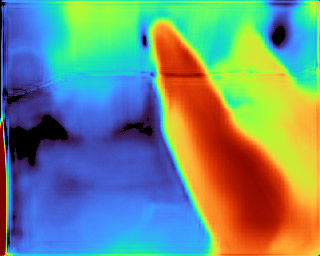} &
\includegraphics[height=\turnheightnew]{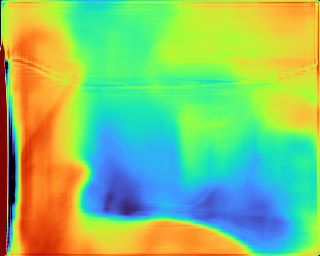} \\

{\rotatebox{90}{\hspace{2mm}{\small DVMVS~\cite{duzceker2021deepvideomvs}}}} &
\includegraphics[height=\turnheightnew]{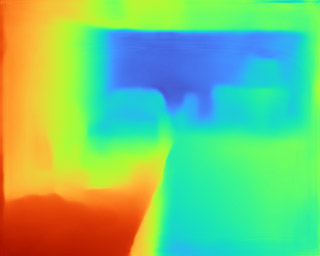} &
\includegraphics[height=\turnheightnew]{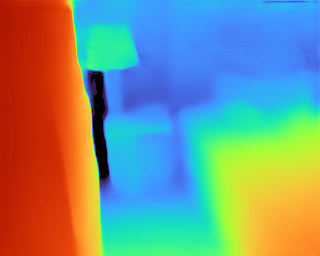} &
\includegraphics[height=\turnheightnew]{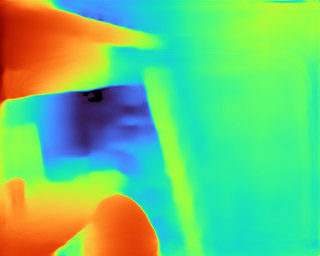} &
\includegraphics[height=\turnheightnew]{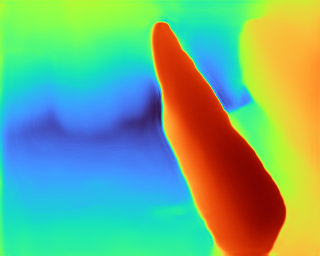} &
\includegraphics[height=\turnheightnew]{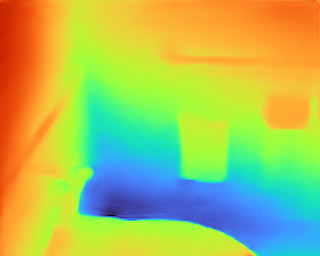} \\

{\rotatebox{90}{\hspace{6mm}{\small \textbf{Ours}}}} &
\includegraphics[height=\turnheightnew]{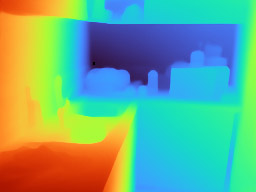} &
\includegraphics[height=\turnheightnew]{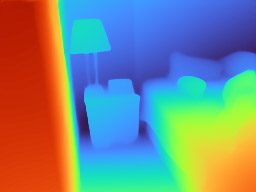} &
\includegraphics[height=\turnheightnew]{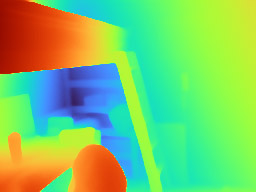} &
\includegraphics[height=\turnheightnew]{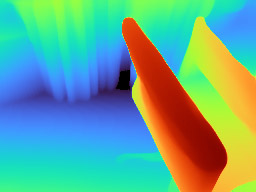} &
\includegraphics[height=\turnheightnew]{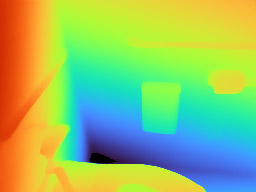} \\

{\rotatebox{90}{\hspace{4mm}{\small GT depth}}} &
\includegraphics[height=\turnheightnew]{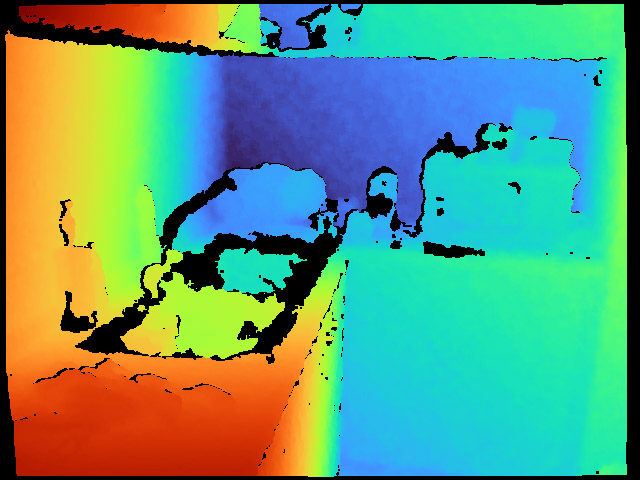} &
\includegraphics[height=\turnheightnew]{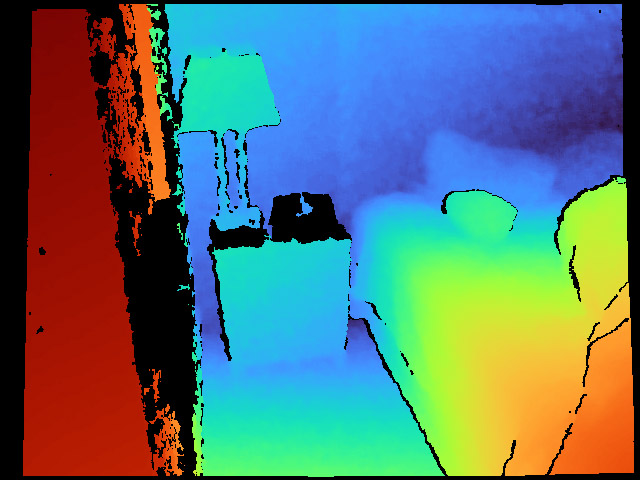} &
\includegraphics[height=\turnheightnew]{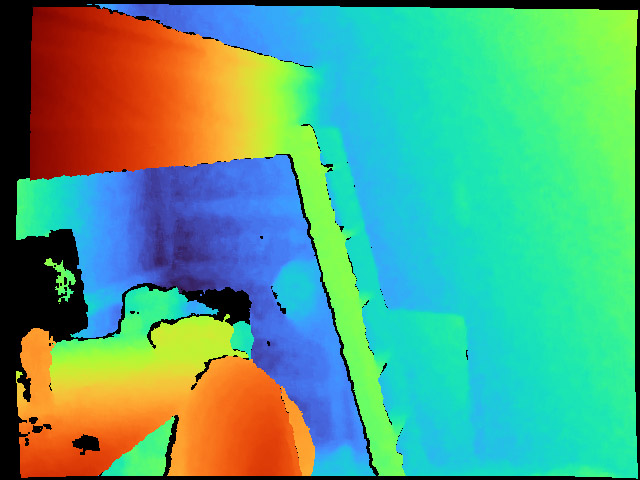} &
\includegraphics[height=\turnheightnew]{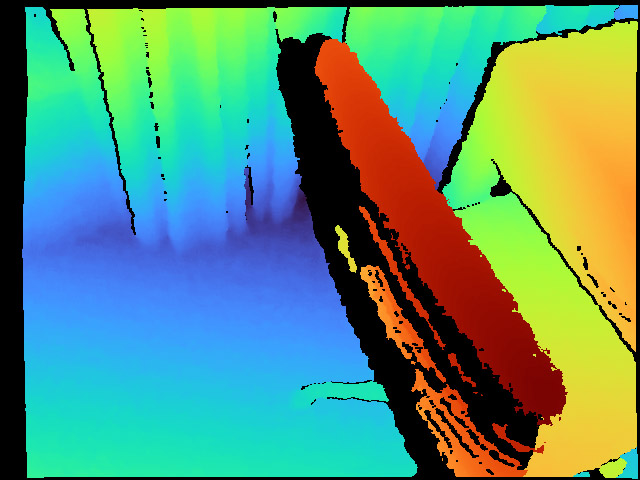} &
\includegraphics[height=\turnheightnew]{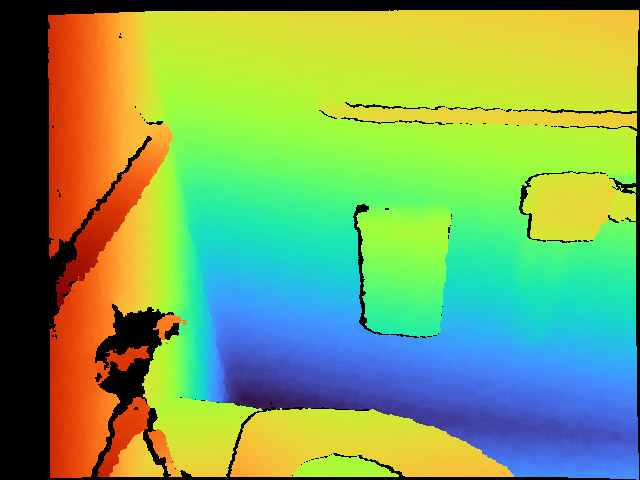} \\

\end{tabular}}
  \caption{\textbf{Depth predictions on ScanNet.} Our model produces significantly sharper and more accurate depths than the baselines. See sup. mat. for additional results.}
  \label{fig:depth_qual_eval}
\end{figure}

\begin{figure}[!h]
  \centering
  \resizebox{\textwidth}{!}{
  \newcommand{\turnheightnew}{0.2\columnwidth}

\newcommand{\rowfive}{scene0757_00_003600}

\centering

\begin{tabular}{@{\hskip -2mm}c@{\hskip 1mm}c@{\hskip 1mm}c@{\hskip 1mm}c@{\hskip 1mm}c@{}}

\includegraphics[height=\turnheightnew]{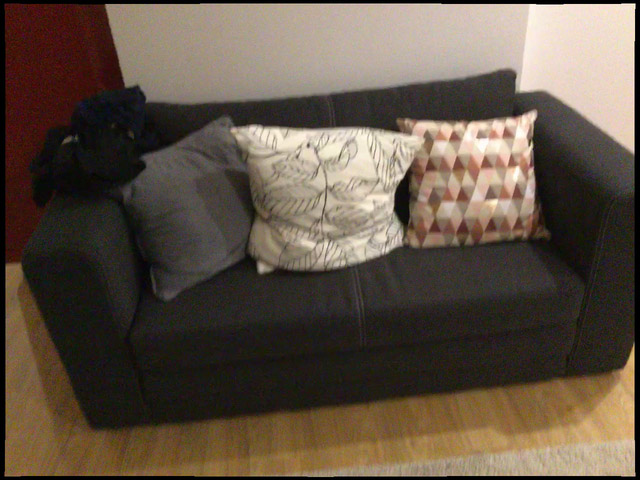} &
\includegraphics[height=\turnheightnew]{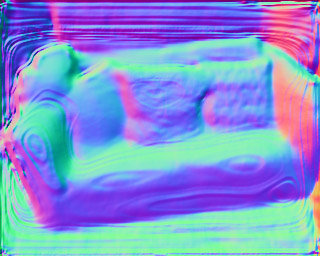} &
\includegraphics[height=\turnheightnew]{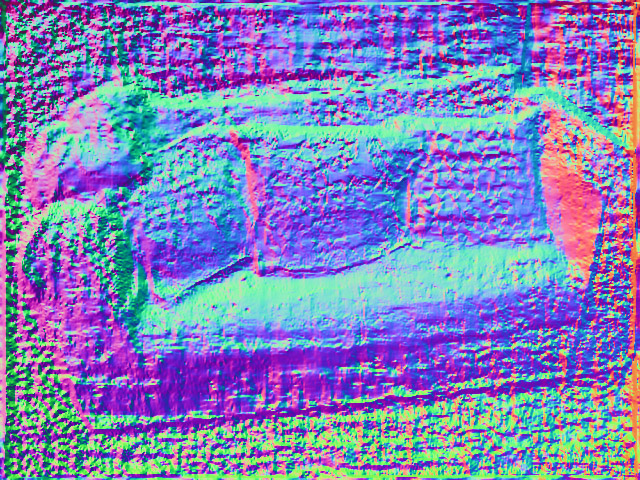} &
\includegraphics[height=\turnheightnew]{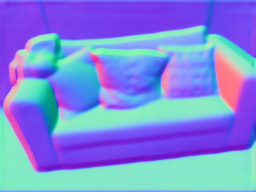} &
\includegraphics[height=\turnheightnew]{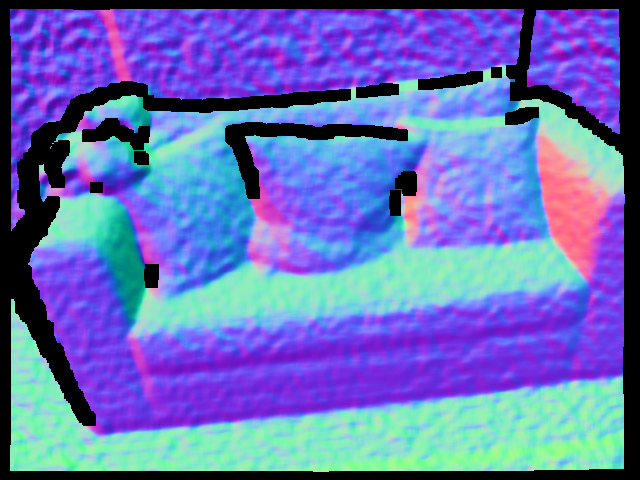} \\

Source Frame &
DVMVS~\cite{duzceker2021deepvideomvs} &
IDNSolver~\cite{zhou2021idn} &
\textbf{Ours} &
GT\\

\end{tabular}}
  \caption{\textbf{Estimated Normals on ScanNet.} Our model produces significantly sharper normals. We compute the estimated normals from depth, see supp. mat. for details.}
  \label{fig:normals_qual_eval}
\end{figure}

We use the standard test split for ScanNetv2 and the test split defined by~\cite{duzceker2021deepvideomvs} for 7-Scenes. We compute depth metrics for every keyframe as in~\cite{duzceker2021deepvideomvs} and average across all keyframes in the test sets.
Surprisingly, our model, which uses no 3D convolutions, outperforms all baselines on depth prediction metrics. 
In addition, our baseline model with no metadata encoding (\ie using only the dot product between reference and source image features) also performs well in comparison to previous methods, showing that a carefully designed and trained 2D network is sufficient for high-quality depth estimation. We show qualitative results for depth and normals in Fig.~\ref{fig:depth_qual_eval} and Fig.~\ref{fig:normals_qual_eval} respectively.

\subsection{3D Reconstruction Evaluation}
Our 3D reconstructions are evaluated using the standard protocol established by TransformerFusion~\cite{bozic2021transformerfusion}.
Their evaluation uses a ground truth mesh based prediction mask to cull away parts of the prediction such that methods are not unfairly penalized for predicting potentially correct geometry that is missing in the ground truth. 
Scores are shown in Table~\ref{table:tf_mesh_evaluation}. 
Our simple depth-based method outperforms state-of-the-art depth estimators for fusion by a wide margin. 
Although we do not perform global refinement of the resulting volume after fusion, we are still able to outperform more expensive \emph{volumetric} methods in some metrics, showing overall competitive performance with lower complexity.

We also compute scores using the ATLAS~\cite{murez2020atlas} mesh evaluation protocol.
However, we find this evaluation is inconsistent; comparing a ground truth mesh against itself does not result in a score of zero, nor does performance on metrics match inspection for visual quality or correlate well across methods. 
These scores, and more details on this discrepancy, are given in the supplementary.


\begin{table*}[tb]
\begin{center}
    \setlength{\tabcolsep}{0.3em} 
    \resizebox{0.85\columnwidth}{!}{  
    \footnotesize
        \begin{tabular}{lcccccccc}
            \toprule
            & Volumetric & Comp$\downarrow$  & Acc$\downarrow$ & Chamfer$\downarrow$ & Prec$\uparrow$ & Recall $\uparrow$ & F-Score $\uparrow$\\
            \midrule
            RevisitingSI~\cite{Hu2018Revisiting} & No & 14.29 & 16.19 & 15.24 & 0.346 & 0.293 & 0.314  \\
            MVDepthNet~\cite{wang2018mvdepthnet}   & No & 12.94 & 8.34 & 10.64 & 0.443 & 0.487 & 0.460  \\
            GPMVS~\cite{hou2019multi}   & No & 12.90 & 8.02 & 10.46 & 0.453 & 0.510 & 0.477  \\
            ESTDepth~\cite{long2021multi} & No & 12.71 & 7.54 & 10.12 & 0.456 & 0.542 & 0.491  \\
            DPSNet~\cite{im2019dpsnet}   & No & 11.94 & 7.58 & 9.77 & 0.474 & 0.519 & 0.492 \\
            DELTAS~\cite{sinha2020deltas} & No & 11.95 & 7.46 & 9.71 & 0.478 & 0.533 & 0.501  \\
            DeepVideoMVS~\cite{duzceker2021deepvideomvs} & No & 10.68 & 6.90 & 8.79 & 0.541 & 0.592 & 0.563  \\
            COLMAP~\cite{schoenberger2016mvs} & No & 10.22 & 11.88 & 11.05 & 0.509 & 0.474 & 0.489 \\
            ATLAS~\cite{murez2020atlas} & Yes & 7.16 & \cellcolor{thirdcolor}7.61 & 7.38 & 0.675 & 0.605 & 0.636 \\
            NeuralRecon~\cite{sun2021neuralrecon} & Yes & \cellcolor{secondcolor}5.09 & 9.13 & 7.11 & 0.630 & \cellcolor{thirdcolor}0.612 & 0.619  \\
            3DVNet~\cite{rich20213dvnet} & Yes & 7.72 & \cellcolor{secondcolor}6.73 & 7.22 & 0.655 & 0.596 & 0.621 \\
            TransformerFusion~\cite{bozic2021transformerfusion} & Yes & \cellcolor{thirdcolor}5.52 & 8.27 & \cellcolor{thirdcolor} 6.89 & \cellcolor{secondcolor}0.728 & 0.600 & \cellcolor{thirdcolor}0.655\\
            VoRTX~\cite{stier2021vortx} & Yes & \cellcolor{firstcolor}4.31 & 7.23 & \cellcolor{firstcolor}5.77 & \cellcolor{firstcolor}0.767 & \cellcolor{secondcolor}0.651 & \cellcolor{firstcolor}0.703  \\
            \textbf{Ours} & No & 5.53 & \cellcolor{firstcolor}6.09 & \cellcolor{secondcolor}5.81 & \cellcolor{thirdcolor}0.686 & \cellcolor{firstcolor}0.658 & \cellcolor{secondcolor}0.671 \\
            \bottomrule
        \end{tabular}
    }
    \end{center}
    \caption{\textbf{Mesh Evaluation}. We use~\cite{bozic2021transformerfusion}'s evaluation. The Volumetric column designates whether a method is a volumetric 3D reconstruction method; other MVS methods that produce only depth maps were reconstructed using standard TSDF fusion. 
    \label{table:tf_mesh_evaluation}}
\end{table*}


\subsection{3D Reconstruction Latency}

For online and interactive 3D reconstruction applications, reducing the latency from sensor reading to 3D representation update is crucial. Most recent reconstruction methods use 3D CNN architectures~\cite{sun2021neuralrecon,murez2020atlas,stier2021vortx,bozic2021transformerfusion} that require expensive and often specialized hardware for sparse matrix computation. This makes them prohibitive for applications on low power devices (smartphones, IoE devices) where both compute and power are limited, or may simply not support the operations. 
Reconstruction methods often report amortized frame time, where the total compute time for select keyframes is averaged over all frames in a sequence. While this is a useful metric for full offline scene reconstruction performance, it is not indicative of online performance, especially when considering latency. 

In Table~\ref{table:timings} we compute the per-frame integration time given a new RGB frame. Some methods may not be designed to run on every keyframe. Notably, NeuralRecon~\cite{sun2021neuralrecon} updates a chunk in world space when 9 keyframes have been received. 
However, for fairness across methods, we do not count the time spent waiting to satisfy a keyframe requirement and assume that the output of immediately available frames with potentially subpar pose distances is comparable to how the method was intended to perform. For methods that require a 3D CNN, we report the time for one 2D keyframe integration and a complete pass of their 3D CNN network.
Although our method is slower than methods such as~\cite{sun2021neuralrecon} on a per-keyframe basis, we can quickly perform updates to the reconstructed volume using online TSDF fusion methods, resulting in low update latencies.

\begin{table*}[tb]
\begin{center}
    \setlength{\tabcolsep}{0.5em} 
    \def\arraystretch{1.2}%
    \resizebox{1.0\columnwidth}{!}{  
        \begin{tabular}{lcccc}
            \toprule
            & \textbf{Volume Update Mode} & \textbf{Breakdown} & \textbf{Update Latency} (ms)$\downarrow$ & \textbf{F-Score}$\uparrow$\\
            \midrule
            ATLAS~\cite{murez2020atlas} & Volume 3D CNN &  2D CNN (29ms) + 3D CNN (353ms) & 382ms & 0.636\\
            NeuralRecon\textbf{*}~\cite{sun2021neuralrecon} & 3D Chunk Fusion + GRU & 2D CNN (12ms) + GRU (78ms) & \cellcolor{secondcolor}90ms & 0.619\\
            3DVNet~\cite{rich20213dvnet} & Iterative 3D CNN & Refine Depths and Feature Cloud (23875ms) & 23875ms & 0.621 \\
            TransformerFusion~\cite{bozic2021transformerfusion} &  Transformer Fusion + 3D CNN  & 2D CNN (131ms) + Refinement (195ms) &  \cellcolor{thirdcolor}326ms & \cellcolor{thirdcolor}0.655\\
            VoRTX~\cite{stier2021vortx} &  Transformer Fusion + 3D CNN &  2D CNN (23ms) + Refinement (4527ms)  & 4550ms & \cellcolor{firstcolor}0.703\\
            \textbf{Ours} & TSDF Fusion & 2D Depth CNN (70ms) + TSDF fuse (2ms) & \cellcolor{firstcolor}72ms & \cellcolor{secondcolor}0.671\\
            
            \bottomrule
        \end{tabular}
    }
    \end{center}
    \caption{\textbf{Frame integration latencies for 3D reconstruction}. We measure latency as the time to incorporate a new image measurement to a 3D representation.  Note that NR reports time amortized over all keyframes. \textbf{*}requires sparse 3D convolutions.}
    \label{table:timings}
\end{table*}

\subsection{Ablations}
\label{sec:ablations}

In order to show the importance of our best practices and novel contributions, we ablate different parts of our network and training routine. Results for depth estimation and mesh reconstruction metrics on ScanNet~\cite{dai2017scannet} are shown for ablations in Table~\ref{table:ablations}, following the evaluation procedures in Section~\ref{sec:experiments}.

\begin{table*}[tb]
\begin{center}
    \setlength{\tabcolsep}{0.3em} 
    \resizebox{0.99\columnwidth}{!}{  
        \begin{tabular}{lccccc|cc}
            & \multicolumn{5}{c}{\textbf{Depth evaluation}} & \multicolumn{2}{c}{\textbf{Mesh eval}} \\
            \cmidrule(lr){2-6}\cmidrule(lr){7-8}
            & Abs Diff$\downarrow$ & Sq Rel$\downarrow$ & RMSE$\downarrow$ & $\delta < 1.05\uparrow$ & $\delta < 1.25\uparrow$ & Chamfer$\downarrow$ & F-score$\uparrow$\\
            \midrule
            \textbf{Ours} w/ all metadata, 8 ordered frames, dot prod CV 16c, ENv2S + R18 & \textbf{0.0885} & \textbf{0.0125} & \textbf{0.1468} & \textbf{73.16} & \textbf{98.09} & \textbf{5.81} & \textbf{67.1} \\
            \midrule
            Ours baseline w/ dot product CV 16c & 0.0941 & 0.0139 & 0.1544 & 70.48 & 97.84 & 6.29 & 64.2 \\
            Ours baseline w/ dot product CV 64c & 0.0944 & 0.0140 & 0.1548 & 70.49 & 97.84 & 6.08 & 65.4 \\
            \midrule
            Ours w/o metadata, shuffled frames & 0.0920 & 0.0135 & 0.1521 & 71.59 & 97.91 & 6.04 & 65.6 \\
            Ours w/ metadata, shuffled frames & 0.0906 & 0.0129 & 0.1490 & 72.09 & 98.03 & 5.92 & 66.3 \\
            \midrule
            Ours baseline w/ dot product CV 16c & 0.0941 & 0.0139 & 0.1544 & 70.48 & 97.84 & 6.29 & 64.2 \\
            Ours dot + feats + mask + depth & 0.0904 & 0.0132 & 0.1509 & 72.63 & 98.03 & 5.92 & 66.5 \\
            Ours dot + feats + mask + depth + ray + angle  & 0.0896 & 0.0127 & 0.1481 & 72.76 & 98.09 &  5.88 & 66.6 \\
            \textbf{Ours} dot + feats + mask + depth + ray + angle + pose distance & \textbf{0.0885} & \textbf{0.0125} & \textbf{0.1468} & \textbf{73.16} & \textbf{98.09} & \textbf{5.81} & \textbf{67.1}\\
            \midrule
            Ours w/ 1 frame -- w/o CV & 0.1742 & 0.0374 & 0.2330 & 40.96 & 90.03 & 9.26 & 47.0 \\
            Ours w/ 2 frames & 0.1230 & 0.0198 & 0.1803 & 57.15 & 96.21 & 7.51 & 56.7 \\
            Ours w/ 4 frames & 0.1036 & 0.0151 & 0.1611 & 65.62 & 97.60 & 6.57 & 62.3 \\
            \midrule\midrule
            Ours w/ metadata but w/ MnasNet at $320\times256$ (matching~\cite{duzceker2021deepvideomvs}) & 0.0947 & 0.0146 & 0.1587 & 71.24 & 97.68 & 5.92 & 66.3 \\
            \bottomrule
        \end{tabular}
    }
    \end{center}
    \caption{\textbf{Ablation Evaluation.} Ablation evaluation on depth and reconstruction metrics using DVMVS keyframes for ScanNet. Scores for our full method are bolded.}
    \label{table:ablations}
\end{table*}

\vspace{5pt}
\noindent\textbf{Baseline ---}
We first show that using \emph{no} MLP and 16 feature channels, reduced using a dot product, our performance greatly suffers.
Interestingly, using 64 feature channels instead of 16 degrades accuracy while being significantly slower. 

\vspace{5pt}
\noindent\textbf{Frame ordering ---} We compare two models where we shuffle the ordering of the keyframes, instead of relying on the pose distance. As we can see while both models suffer from random ordering, the full model, which has access to the pose distance as metadata, does not suffer as much.

\vspace{5pt}
\noindent\textbf{Metadata ---} In this section all the models make use of the MLP cost volume reduction, but we vary the input of that MLP. We start with our baseline model, using only the feature dot products aggregated using a sum. We then add the features, their depth and validity mask, reduced using our MLP. We keep adding more metadata until we reach our full model. Accuracy increases with the amount of information provided to the model.

\vspace{5pt}
\noindent\textbf{Views ---}
In addition, we show that our method can incorporate information from many source views. As we increase from 2 views to 8, our performance continues to improve. In contrast, DeepVideoMVS's performance remains relatively constant when using more than three source frames~\cite{duzceker2021deepvideomvs}. In addition, we ablate the cost volume entirely by zeroing its output (creating a monocular method), leading to greatly decreased performance, showing that a strong metric depth estimate from the cost volume is required to resolve scale ambiguity. 

\section{Conclusion}

We propose SimpleRecon, which produces state-of-the-art depth estimations and 3D reconstructions, all without the use of expensive 3D convolutions.
Our key contribution is to inject cheaply available \emph{metadata} into the cost volume.
Our evaluation shows that metadata boosts scores, and in partnership with our set of careful architecture design choices leads to our state-of-the-art depths.
Moreover, our method does not preclude the use of 3D convolutions or additional cost volume and depth refinement techniques, allowing room for further improvements when compute is less restricted. 
Ultimately, our back-to-basics approach shows that high-quality depths are all you need for high-quality reconstructions.


\paragraph{Acknowledgments} ---
We thank Alja\v{z} Bo\v{z}i\v{c} \cite{bozic2021transformerfusion}, Jiaming Sun \cite{sun2021neuralrecon} and Arda D{\"u}z\c{c}eker \cite{duzceker2021deepvideomvs} for quickly providing useful information to help with baselines.
Mohamed is funded by a Microsoft Research PhD Scholarship (MRL 2018-085).


\clearpage
%
%
\bibliographystyle{splncs04}
\bibliography{bib}
\end{document}